%% file: main.tex
\documentclass{article}

\usepackage{microtype}
\usepackage{graphicx}
\usepackage{subcaption}
\usepackage{booktabs} 

\usepackage{hyperref}

\usepackage[preprint]{icml2026}

\usepackage{amsmath}
\usepackage{amssymb}
\usepackage{mathtools}
\usepackage{amsthm}

\usepackage[capitalize,noabbrev]{cleveref}

\theoremstyle{plain}

\theoremstyle{definition}

\theoremstyle{remark}

\usepackage[textsize=tiny]{todonotes}
\definecolor{cvprblue}{rgb}{0.21,0.49,0.74}
\usepackage[table]{xcolor}
\usepackage{multirow}
\usepackage{pifont}

\icmltitlerunning{Towards a Unified Driving World Model via Multifaceted Representation Learning}

\begin{document}

\twocolumn[
  \icmltitle{UniDWM: Towards a Unified Driving World Model via Multifaceted Representation Learning}



  \icmlsetsymbol{equal}{*}

  \begin{icmlauthorlist}
    \icmlauthor{Shuai Liu}{sysu}
    \icmlauthor{Siheng Ren}{sysu}
    \icmlauthor{Xiaoyao Zhu}{sysu}
    \icmlauthor{Quanmin Liang}{sysu}\\
    \icmlauthor{Zefeng Li}{sysu}
    \icmlauthor{Qiang Li}{xpeng}
    \icmlauthor{Xin Hu}{xpeng}
    \icmlauthor{Kai Huang}{sysu}
  \end{icmlauthorlist}

  \icmlaffiliation{sysu}{School of Computer Science and Engineering, Sun Yat-sen University}
  \icmlaffiliation{xpeng}{Xpeng Motors Technology Co Ltd}

  \icmlcorrespondingauthor{Kai Huang}{huangk36@mail.sysu.edu.cn}
   \icmlcorrespondingauthor{Shuai Liu}{liush376@mail2.sysu.edu.cn}
  \icmlkeywords{Machine Learning, ICML}

  \vskip 0.3in
]



\printAffiliationsAndNotice{}  

\input{sec/0_abstract}    
\input{sec/1_intro}

\input{sec/2_related}
\input{sec/3_method}
\input{sec/4_exp}

\section{Conclusion}
We presented UniDWM, a unified driving world model that learns a structure- and dynamics-aware latent representation as a physically grounded state space for autonomous driving. UniDWM jointly reconstructs scene appearance, geometry, and ego-motion from visual observations, and models future world evolution via a diffusion-based generative process, enabling coherent reasoning across perception, prediction, and planning within a single latent space. By formulating UniDWM as a variant of the variational autoencoder, we provide principled theoretical grounding for multifaceted representation learning beyond heuristic design choices. Extensive evaluations on the NAVSIM benchmark demonstrate that UniDWM consistently improves trajectory planning while achieving strong performance in 4D reconstruction and generation, validating the effectiveness of unified world modeling for planning-oriented autonomous driving.

\section*{Impact Statement}
This paper presents work whose goal is to facilitate the learning and applications of world models. There are many potential societal consequences of our work, none of which we feel must be specifically highlighted here.


\bibliography{main}
\bibliographystyle{icml2026}

\newpage
\appendix
\onecolumn

\section{Proof of VAE-style ELBO Objective of UniDWM}
\label{vae_unidwm}
\subsection{Problem Setup}

We consider a dataset
\[
\mathcal{D} = \{ \mathbf{x}_i \in \mathcal{X} \}_{i=1}^{N},
\]
where each scene $\mathbf{x}_i$ consists of $M$ observations from different perspectives,
\[
\mathbf{x}_i = \{ \boldsymbol{x}_i^{(j)} \}_{j=1}^{M}.
\]
Each observation $\boldsymbol{x}^{(j)}$ corresponds to a single frame--modality pair.
We assume that scenes are i.i.d.\ samples from an unknown data distribution
$p_{\mathcal{D}}(\mathbf{x})$.

Given a subset of local observations $\mathbf{x}^{local} \subset \mathbf{x}$,
the objective of UniDWM is to reconstruct the full set of global observations
$\mathbf{x}$.
Let $\mathbf{z} \in \mathcal{Z}$ denote a latent variable capturing the global
scene representation.

The generative model is defined as
\[
p_\theta(\mathbf{x}, \mathbf{z})
= p(\mathbf{z}) \prod_{j=1}^{M} p_\theta(\boldsymbol{x}^{(j)} \mid \mathbf{z}),
\]
where conditional independence of observations given $\mathbf{z}$ is assumed.
The approximate posterior is given by
\[
q_\phi(\mathbf{z} \mid \mathbf{x}^{local}).
\]

\subsection{Variational Lower Bound}

We begin with the log-likelihood of the full observations:
\[
\log p_\theta(\mathbf{x})
= \log \int p_\theta(\mathbf{x}, \mathbf{z})
\, d\mathbf{z}.
\]

Introducing the variational posterior $q_\phi(\mathbf{z} \mid \mathbf{x}^{local})$,
we rewrite the marginal likelihood as
\[
\log p_\theta(\mathbf{x})
= \log \int q_\phi(\mathbf{z} \mid \mathbf{x}^{local})
\frac{p_\theta(\mathbf{x}, \mathbf{z})}
     {q_\phi(\mathbf{z} \mid \mathbf{x}^{local})}
\, d\mathbf{z}.
\]

Applying Jensen's inequality yields the following lower bound:
\[
\log p_\theta(\mathbf{x})
\ge
\mathbb{E}_{q_\phi(\mathbf{z} \mid \mathbf{x}^{local})}
\left[
\log \frac{p_\theta(\mathbf{x}, \mathbf{z})}
           {q_\phi(\mathbf{z} \mid \mathbf{x}^{local})}
\right].
\]

Substituting the factorized joint distribution, we obtain
\[
\begin{aligned}
\log p_\theta(\mathbf{x})
\ge\;&
\mathbb{E}_{q_\phi(\mathbf{z} \mid \mathbf{x}^{local})}
\Bigg[
\sum_{j=1}^{M} \log p_\theta(\boldsymbol{x}^{(j)} \mid \mathbf{z})
+ \log p(\mathbf{z}) \\
&\qquad\qquad - \log q_\phi(\mathbf{z} \mid \mathbf{x}^{local})
\Bigg].
\end{aligned}
\]

Rearranging terms gives
\[
\begin{aligned}
\log p_\theta(\mathbf{x})
\ge\;&
\sum_{j=1}^{M}
\mathbb{E}_{q_\phi(\mathbf{z} \mid \mathbf{x}^{local})}
\big[ \log p_\theta(\boldsymbol{x}^{(j)} \mid \mathbf{z}) \big] \\
&-
\mathrm{KL}\!\left(
q_\phi(\mathbf{z} \mid \mathbf{x}^{local})
\,\|\, p(\mathbf{z})
\right).
\end{aligned}
\]

\subsection{Dataset-Level Objective}

Taking the expectation over the data distribution $p_{\mathcal{D}}(\mathbf{x})$,
we arrive at the dataset-level ELBO:
\[
\begin{aligned}
\mathcal{L}_{\mathrm{UniDWM0}} \equiv\;&
{\textstyle \sum_{j=1}^{M}}
\mathbb{E}_{p_{\mathcal{D}}(\mathbf{x})}
\mathbb{E}_{q_\phi(\mathbf{z} \mid \mathbf{x}^{local})}
\big[ \log p_\theta(\boldsymbol{x}^{(j)} \mid \mathbf{z}) \big] \\
&-
\mathbb{E}_{p_{\mathcal{D}}(\mathbf{x})}
\mathrm{KL}\!\left(
q_\phi(\mathbf{z} \mid \mathbf{x}^{local})
\,\|\, p(\mathbf{z})
\right),
\end{aligned}
\]
which matches Eq.~\ref{Eq:unidwm0}.

\section{Proof of the InfoVAE-style Objective for UniDWM}
\label{infovae_unidwm}

In this section, we derive the InfoVAE-inspired objective used in UniDWM and
show how it follows from a decomposition of the standard VAE ELBO.

\subsection{Background: Limitations of the VAE ELBO}

Recall the standard ELBO derived in Appendix~\ref{vae_unidwm}:
\[
\begin{aligned}
\mathcal{L}_{\mathrm{UniDWM0}} =
{\textstyle \sum_{j=1}^{M}}
\mathbb{E}_{p_{\mathcal{D}}(\mathbf{x})}
\mathbb{E}_{q_\phi(\mathbf{z} \mid \mathbf{x}^{local})}
\big[ \log p_\theta(\boldsymbol{x}^{(j)} \mid \mathbf{z}) \big]
-
\mathbb{E}_{p_{\mathcal{D}}(\mathbf{x})}
\mathrm{KL}\!\left(
q_\phi(\mathbf{z} \mid \mathbf{x}^{local})
\,\|\, p(\mathbf{z})
\right).
\end{aligned}
\]

When the observation space $\mathcal{X}$ is complex and high-dimensional,
the KL regularization term may encourage the approximate posterior
$q_\phi(\mathbf{z} \mid \mathbf{x}^{local})$ to collapse toward the prior
$p(\mathbf{z})$, resulting in uninformative latent representations.
This phenomenon is commonly referred to as \emph{posterior collapse}.

To alleviate this issue, InfoVAE~\cite{zhao2017infovae} proposes an alternative
regularization that explicitly controls the \emph{aggregated posterior}
$q_\phi(\mathbf{z})$ rather than the conditional posterior.

\subsection{Decomposition of the KL Term}

We define the aggregated posterior as
\[
q_\phi(\mathbf{z})
\triangleq
\mathbb{E}_{p_{\mathcal{D}}(\mathbf{x})}
\big[ q_\phi(\mathbf{z} \mid \mathbf{x}^{local}) \big],
\]
where $\mathbf{x}^{local}$ is obtained from $\mathbf{x}$ through a fixed and
deterministic sampling strategy.

The expected KL divergence in the standard ELBO can be decomposed as
\[
\begin{aligned}
\mathbb{E}_{p_{\mathcal{D}}(\mathbf{x})}
\mathrm{KL}\!\left(
q_\phi(\mathbf{z} \mid \mathbf{x}^{local})
\,\|\, p(\mathbf{z})
\right)
=&\;
I_{q}(\mathbf{z}; \mathbf{x}^{local})
+
\mathrm{KL}\!\left(
q_\phi(\mathbf{z}) \,\|\, p(\mathbf{z})
\right),
\end{aligned}
\]
where
\[
I_{q}(\mathbf{z}; \mathbf{x}^{local})
\triangleq
\mathbb{E}_{p_{\mathcal{D}}(\mathbf{x})}
\mathrm{KL}\!\left(
q_\phi(\mathbf{z} \mid \mathbf{x}^{local})
\,\|\, q_\phi(\mathbf{z})
\right)
\]
is the mutual information between $\mathbf{z}$ and $\mathbf{x}^{local}$ under
the variational joint distribution.

Substituting this decomposition into the ELBO yields
\[
\begin{aligned}
\mathcal{L}_{\mathrm{UniDWM0}}
=&\;
{\textstyle \sum_{j=1}^{M}}
\mathbb{E}_{p_{\mathcal{D}}(\mathbf{x})}
\mathbb{E}_{q_\phi(\mathbf{z} \mid \mathbf{x}^{local})}
\big[ \log p_\theta(\boldsymbol{x}^{(j)} \mid \mathbf{z}) \big] \\
&-
I_{q}(\mathbf{z}; \mathbf{x}^{local})
-
\mathrm{KL}\!\left(
q_\phi(\mathbf{z}) \,\|\, p(\mathbf{z})
\right).
\end{aligned}
\]

This form reveals that the standard ELBO implicitly penalizes the mutual
information $I_q(\mathbf{z}; \mathbf{x}^{local})$, which discourages the latent
variable from encoding meaningful information about the observations.

\subsection{InfoVAE-style Objective}

InfoVAE~\cite{zhao2017infovae} introduces a generalized objective that allows
explicit control over the relative importance of the mutual information term
and the aggregated posterior regularization. The general InfoVAE objective can
be written equivalently as
\[
\begin{aligned}
\mathcal{L}_{\mathrm{InfoVAE}}
=&\;
\mathbb{E}_{p_{\mathcal{D}}(\mathbf{x})}
\mathbb{E}_{q_\phi(\mathbf{z} \mid \mathbf{x})}
\big[ \log p_\theta(\mathbf{x} \mid \mathbf{z}) \big] \\
&-
(1-\alpha)\, I_q(\mathbf{z}; \mathbf{x}) 
- \lambda D\!\left(
q_\phi(\mathbf{z}) \,\|\, p(\mathbf{z})
\right),
\end{aligned}
\]
where $D(\cdot\|\cdot)$ is a general divergence satisfying
$D(q_\phi(\mathbf{z})\|p(\mathbf{z}))=0$ if and only if
$q_\phi(\mathbf{z})=p(\mathbf{z})$, $\alpha$ and $\lambda$ are scaling coefficient. 

In UniDWM, we adopt the InfoVAE formulation and set $\alpha = 1$.
As a result, the mutual information penalty is completely removed, and the
objective reduces to
\[
\begin{aligned}
\mathcal{L}_{\mathrm{UniDWM}}
=&\;
{\textstyle \sum_{j=1}^{M}}
\mathbb{E}_{p_{\mathcal{D}}(\mathbf{x})}
\mathbb{E}_{q_\phi(\mathbf{z} \mid \mathbf{x}^{local})}
\big[ \log p_\theta(\boldsymbol{x}^{(j)} \mid \mathbf{z}) \big] \\
&-
\lambda\,
D\!\left(
q_\phi(\mathbf{z}) \,\|\, p(\mathbf{z})
\right).
\end{aligned}
\]
Therefore, the UniDWM objective can be viewed as a direct instantiation of
InfoVAE with $\alpha = 1$, extended to the multi-observation reconstruction
setting.

\section{Dataset and Metrics}
\label{dataset_metrics}
We mainly evaluate models on the NAVSIM \cite{navsim} dataset, which provides 120 hours of challenging driving data with high-resolution camera images and LiDAR inputs spanning up to 1.5 seconds. It filters out trivial driving scenarios to emphasize complex decision-making. For the trajectory planning task, we use the official Predictive Driver Model Score (PDMS) as the evaluation metric. For the 3D reconstruction task, following prior works \cite{wang2024dust3r, vggt}, we report Accuracy (the smallest Euclidean distance from the prediction to ground truth), Completeness (the smallest Euclidean distance from the ground truth to prediction), and their average Overall (i.e., Chamfer distance).

\section{Smoothness Analysis of Learned Representations}
\label{sec:appendix_smoothness}

We perform a smoothness analysis of the latent representations obtained from the baseline model (the same as that in Table~\ref{tab:grpo}) and our UniDWM on the NAVSIM \textit{test} split. The goal of this analysis is to assess how smooth the learned representations are and to investigate potential correlations with robustness and generalization in downstream planning tasks. For each model, we extract latent representations for all samples in the test split and apply feature-wise standardization. We then construct local neighborhoods using $k$-nearest neighbors ($k=32$ in practice) in the representation space and compute the following complementary smoothness metrics:

\begin{itemize}
    \item \textbf{kNN average distance} (kNN Dist.): average distance to $k$ nearest neighbors, capturing local sensitivity in the latent space.
    \item \textbf{Local PCA energy ratio} (PCA Ratio): fraction of variance explained by the top principal components within each local neighborhood, reflecting local linearity.
    \item \textbf{Graph Laplacian smoothness} (Lap. Smooth.): graph Laplacian smoothness over the $k$-nearest neighbor graph, measuring consistency across the latent manifold.
\end{itemize}

The computed smoothness metrics for the baseline model and UniDWM are summarized in Table~\ref{tab:smoothness_results}. Compared to the baseline, UniDWM achieves: 1) Lower kNN average distance, indicating that neighboring samples are closer in the latent space and local variations are smaller. 2) Higher local PCA energy ratio, suggesting that local neighborhoods are better approximated by low-dimensional linear subspaces. 3) Lower Graph Laplacian smoothness, reflecting smoother variation across the latent manifold. These results consistently indicate that UniDWM produces smoother latent representations than the baseline, which is expected to contribute to more stable latent rollouts and improved robustness and generalization in downstream tasks.

\begin{table}[h!]
\centering
\caption{Smoothness metrics of latent representations on the NAVSIM \textit{test} split. Lower kNN distance and Laplacian smoothness, and higher PCA ratio, indicate smoother representations.}
\label{tab:smoothness_results}
\begin{tabular}{lccc}
\toprule
Model & kNN Dist. & PCA Ratio & Lap. Smooth. \\
\midrule
Baseline & 130.578 & 0.554 & 328,596 \\
UniDWM   & 117.238 & 0.622 & 263,981 \\
\bottomrule
\end{tabular}
\end{table}

\section{Limitations} 
Our framework is primarily trained and evaluated on sequential monocular visual inputs, and does not explicitly explore multi-view visual or multi-sensory inputs such as LiDAR and radar. Second, although the diffusion-based generation module enables expressive and flexible modeling of future world dynamics, it introduces non-trivial computational overhead during training. Third, UniDWM currently models future evolution within a limited temporal horizon (next-frame prediction in practice). We leave the exploration of multimodal, lightweight, and long-horizon frameworks to future work.



\end{document}

%% file: sec/0_abstract.tex
\begin{abstract}
Achieving reliable and efficient planning in complex driving environments requires a model that can reason over the scene’s geometry, appearance, and dynamics. We present \textbf{UniDWM}, a \textbf{unified driving world model} that advances autonomous driving through \textit{multifaceted representation learning}. UniDWM constructs a structure- and dynamic-aware latent world representation that serves as a physically grounded state space, enabling consistent reasoning across perception, prediction, and planning. Specifically, a \textit{joint reconstruction} pathway learns to recover the scene’s structure, including geometry and visual texture, while a \textit{collaborative generation} framework leverages a conditional diffusion transformer to forecast future world evolution within the latent space. Furthermore, we show that our UniDWM can be deemed as a variation of VAE, which provides theoretical guidance for the multifaceted representation learning. Extensive experiments demonstrate the effectiveness of UniDWM in trajectory planning, 4D reconstruction and generation, highlighting the potential of multifaceted world representations as a foundation for unified driving intelligence. The code will be publicly available at \url{https://github.com/Say2L/UniDWM}.
\end{abstract}

%% file: sec/1_intro.tex
\section{Introduction}
\label{sec:intro}

Reliable planning in autonomous driving hinges on a comprehensive understanding of the physical world—one that captures not only the visual appearance of a scene, but also its underlying geometry and dynamic evolution over time. Recent advances in autonomous driving \cite{hu2023planning, jiang2023vad, chen2024vadv2, liao2025diffusiondrive, liu2025gaussianfusion} have made substantial progress by reformulating the traditional perception–prediction–planning pipeline into an end-to-end paradigm, thereby mitigating error accumulation across stages. To learn representations that jointly encode appearance, geometry, and scene dynamics, these approaches, however, still heavily depend on costly perception annotations, which limit the scalability.

Humans make driving decisions by mentally simulating the evolving world—anticipating the future states of surrounding agents, objects, and the ego vehicle according to limited observations. This process implicitly builds a coherent world model that tightly couples visual perception, spatial reasoning, and action planning within a unified framework. Motivated by this observation, we posit that autonomous agents should be equipped with a unified driving world model capable of jointly reconstructing and forecasting the 4D world state from partial observations, such as multi-view visual inputs. Such a formulation naturally lends itself to self-supervised learning, where latent world representations inferred from limited sensory inputs can be leveraged to recover unobserved information, including scene geometry and future world dynamics.

Learning a unified driving world model solely from visual observations presents several fundamental challenges. First, establishing \textbf{\emph{spatio--temporal coherence}} is crucial for modeling realistic dynamics. The latent representation must not only encode per-frame spatial structures but also preserve temporal continuity, faithfully reflecting how scenes and agents evolve over time under causal motion. Second, there exists a \textbf{\emph{lack of theoretical guidance}} for learning unified driving world models. Most existing methods \cite{epona, vggt, liu2025gaussianfusion} are based on heuristic designs, with limited understanding of which principles are essential for jointly modeling the dynamic physical world, thereby hindering systematic improvement and generalization.

To tackle these challenges, we propose \textbf{UniDWM}, a unified driving world model that learns a structure- and dynamic-aware latent world representation serving as a physically grounded state space. This representation enables coherent reasoning across perception, prediction, and planning. As illustrated in Figure~\ref{fig_framework}, UniDWM comprises two key components. First, a \textit{Joint Reconstruction} pathway reconstructs the scene structure---including geometry, visual appearance, and ego-motion---from sequential observations, yielding a unified latent world representation. Second, a \textit{Collaborative Generation} module leverages a conditional diffusion transformer (DiT) to predict future world evolution, thereby implicitly injecting dynamics-aware information into the latent space. Furthermore, we demonstrate that UniDWM can be formulated as a variant of a variational autoencoder (VAE)~\cite{kingma2014auto, zhao2017infovae}, providing theoretical grounding for its unified representation learning paradigm.

Through comprehensive evaluations on NAVSIM, spanning trajectory planning, 4D reconstruction, and 4D generation, we show that UniDWM effectively unifies perception, prediction, and planning within a single latent space, yielding tangible benefits for planning. Our main contributions are summarized as follows:
\begin{itemize}
    \setlength{\itemsep}{0pt}
    \setlength{\parskip}{0pt}
    \setlength{\parsep}{0pt}
    \item We propose UniDWM, a unified driving world model that learns a structure- and dynamic-aware latent representation as a physically grounded state space for perception, prediction, and planning.
    \item We provide a theoretical interpretation of UniDWM as a variant of a variational autoencoder, offering principled guidance for multifaceted representation learning.
    \item We evaluate UniDWM extensively on the NAVSIM benchmark across trajectory planning, 4D reconstruction, and 4D generation, demonstrating the effectiveness of our method.
\end{itemize}

%% file: sec/2_related.tex
\section{Related Work}
\label{sec:related}

\noindent \textbf{Driving World Models.}
Recent advances in world models have steered autonomous driving research toward data-driven generative reasoning. Existing approaches can be broadly divided into two categories. The first line of work focuses on scene-level generation for simulation and data augmentation. Methods such as VISTA~\cite{GaoVista2024} and GAIA-2~\cite{RussellGAIA22025} enable controllable scene synthesis via viewpoint- and semantics-conditioned diffusion, while MagicDrive-V2~\cite{magicdrivev2} and Genesis~\cite{guo2025genesis} further incorporate structured layouts to produce realistic driving scenarios. The second category investigates world-model-based driving control~\cite{epona,zheng2025world4drive,shi2025drivex,wote}, where learned dynamics and generative predictions are integrated into decision-making processes. Specifically, Epona~\cite{epona} combines diffusion-based prediction with end-to-end planning, whereas DriveX~\cite{shi2025drivex} and WoTe~\cite{wote} embed semantic information into BEV latent representations. In contrast, our UniDWM focuses on learning a unified latent world representation that jointly models spatial structure and dynamics, thereby more closely approximating the true underlying world and improving generalization across downstream tasks.

\noindent\textbf{End-to-End Trajectory Planning.}
End-to-end autonomous driving (E2E-AD) directly maps sensor inputs to control outputs, bypassing modular perception–prediction-planning pipelines. Early CNN-based methods~\cite{bojarski2016end} predicted steering from camera images, later enhanced by conditional imitation learning (CIL)~\cite{codevilla2018end} to incorporate high-level navigation commands. CILRS~\cite{codevilla2019exploring} improved generalization with residual architectures and auxiliary targets. More recent methods explore joint optimization and structured representations. UniAD~\cite{hu2023planning} employs a query-based BEV design for joint optimization, while VAD~\cite{jiang2023vad} and VADv2~\cite{chen2024vadv2} employ vectorized representations to model the driving scene for efficiency. Generative methods, such as diffusion-conditioned trajectory generators~\cite{xing2025goalflow, liao2025diffusiondrive} and momentum-aware planning~\cite{song2025don}, further enhance robustness and uncertainty handling. Despite these advances, most E2E methods still rely heavily on costly perception annotations, as supervision based solely on trajectories often leads to representation collapse or shortcut learning behaviors~\cite{transfuser, navsim}. In contrast, UniDWM adopts a self-supervised learning paradigm to learn a unified latent world representation, which is encouraged to be smooth and well-structured through multiple reconstruction and regularization objectives.

\noindent\textbf{Representation Learning for Autonomous Driving.}
Learning an effective world representation is essential for reasoning about perception, prediction, and planning. BEV-based methods~\cite{li2024bevformer, liang2022bevfusion, jia2023think} provide structured spatial alignment, enabling joint detection, mapping, and motion prediction, but suffer from information loss from projection. Image-centric multi-view representations~\cite{wang2022detr3d, liu2022petr, jia2025drivetransformer} preserve perspective cues for fine-grained perception but introduce depth ambiguity. Gaussian-based scene representations~\cite{3dgs, kerbl2024hierarchical, huang2024gaussianformer, liu2025gaussianfusion} model geometry and appearance continuously with real-time rendering, though they are often optimized offline. Overall, BEV methods rely on strong inductive biases, image-centric approaches lack explicit geometric modeling, and Gaussian-based representations are rarely explored in the context of trajectory planning. UniDWM aims to learn a unified latent world representation with strong generalization ability that can support multiple downstream tasks with minimal inductive bias. UniDWM achieves this goal through multifaceted representation learning that jointly captures geometry, appearance, and evolution dynamics, thereby providing a foundation for physically grounded world modeling.

%% file: sec/3_method.tex
\begin{figure*}[t]
    \centering
    \includegraphics[width=0.9\textwidth]{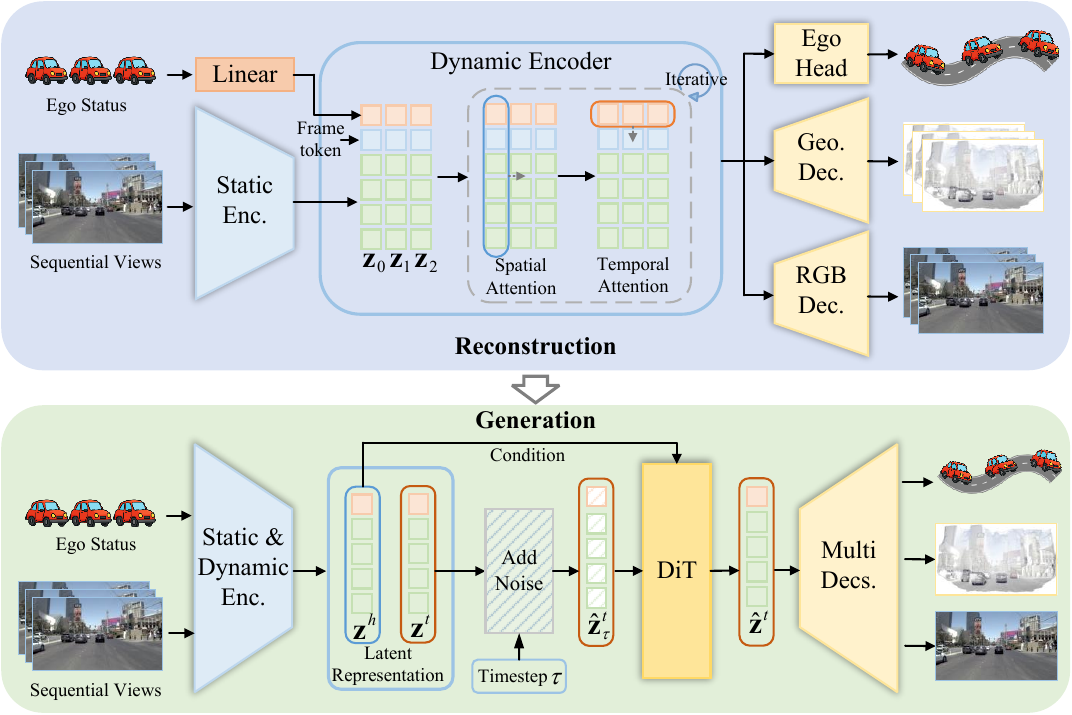}
    \caption{\textbf{The overall framework of our UniDWM.} UniDWM is composed of the Reconstruction and Generation stages. Given a sequence of visual observations as input, UniDWM projects them to a unified feature space using static and dynamic encoders. Then, reconstruction and generation decoders can be used to reconstruct the observed scene and generate dynamic evolution, respectively. }
    \label{fig_framework}
\end{figure*}

\section{Method}

\subsection{Preliminary}

Variational Autoencoder (VAE)~\cite{kingma2014auto} is a probabilistic generative framework that introduces latent variables
$z \in \mathcal{Z}$ to explain observations $\mathbf{x} \in \mathcal{X}$.
The model assumes a simple prior distribution $p(\mathbf{z})$ over the latent space, commonly chosen as a Gaussian distribution,
and defines a conditional data distribution $p_\theta(\mathbf{x} | \mathbf{z})$ parameterized by a neural network.
Given the data-generating distribution $p_{\mathcal{D}}(\mathbf{x})$, approximated by a finite training set,
learning is naturally formulated as maximizing the expected marginal log-likelihood:
\begin{equation}
\mathbb{E}_{p_{\mathcal{D}}(\mathbf{x})} \big[ \log p_\theta(\mathbf{x}) \big]
=
\mathbb{E}_{p_{\mathcal{D}}(\mathbf{x})} \Big[ \log \mathbb{E}_{p(\mathbf{z})} \big[ p_\theta(\mathbf{x} \mid \mathbf{z}) \big] \Big].
\label{Eq:vae_likelihood}
\end{equation}
Directly optimizing this objective is generally infeasible due to the intractable integration over latent variables.
To overcome this difficulty, VAE introduces a variational posterior $q_\phi(\mathbf{z} | \mathbf{x})$
and maximizes a tractable lower bound of the log-likelihood, given by:
\begin{equation}
\mathcal{L}_{\text{VAE}} =
\mathbb{E}_{q_\phi(\mathbf{z} \mid \mathbf{x})} \big[ \log p_\theta(\mathbf{x} \mid \mathbf{z}) \big]
-
\mathrm{KL}\!\left(q_\phi(\mathbf{z} \mid \mathbf{x}) \,\|\, p(\mathbf{z})\right),
\label{Eq:vae}
\end{equation}
which satisfies $\mathcal{L}_{\text{VAE}} \le \log p_\theta(\mathbf{x})$.

While effective in practice, the standard VAE objective implicitly limits the mutual information between observations and latent variables,
often resulting in weakly informative representations.
InfoVAE~\cite{zhao2017infovae} alleviates this issue by decoupling information preservation from prior regularization.
The resulting objective is formulated as: 
\begin{equation}
\begin{aligned}
\mathcal{L}_{\text{InfoVAE}} \equiv\;&
\mathbb{E}_{p_{\mathcal{D}}(\mathbf{x})}
\mathbb{E}_{q_\phi(\mathbf{z} \mid \mathbf{x})}
\big[ \log p_\theta(\mathbf{x} \mid \mathbf{z}) \big] \\
&- (1 - \alpha)\,
\mathbb{E}_{p_{\mathcal{D}}(\mathbf{x})}
\mathrm{KL}\!\left(q_\phi(\mathbf{z} \mid \mathbf{x}) \,\|\, p(\mathbf{z})\right) \\
&- (\alpha + \lambda - 1)\,
\mathrm{D}\!\left(q_\phi(\mathbf{z}) \,\|\, p(\mathbf{z})\right),
\end{aligned}
\label{Eq:infovae}
\end{equation}
where $q_\phi(z) = \mathbb{E}_{p_{\mathcal{D}}(x)} q_\phi(z | x)$ denotes the aggregated posterior.
Here, $D(\cdot\|\cdot)$ represents a divergence measure such as KL divergence or maximum mean discrepancy, and the hyperparameters $\alpha$ and $\lambda$ balance reconstruction fidelity, prior matching, and information preservation.

\subsection{Unified Driving World Model}
We consider a driving dataset $\mathcal{D} = \{ \mathbf{x}_i \in \mathcal{X}\}_{i=1}^{N}$ of $N$ i.i.d. scenes, each of which is a set of $M$ observations $\mathbf{x}_i = \{\boldsymbol{x}_i^{(j)}\}_{j=1}^{M}$ from different perspectives. We consider a frame with a single modality as an observation perspective. If a scene in the dataset is composed of $l$ frames and $m$ modalities (e.g., points, images, ego poses, etc.), then $M$ is equal to $l \times m$. Given some local observations $\mathbf{x}^{local} \in \mathbf{x}$ of a scene, the objective of UniDWM is reconstruct the global observations $\mathbf{x}$. The ELBO of UniDWM can be written as:
\begin{equation}
\begin{aligned}
\mathcal{L}_{\text{UniDWM0}} \equiv\;&
 {\textstyle \sum_{j=1}^{M}} \mathbb{E}_{p_{\mathcal{D}}(\mathbf{x})}
\mathbb{E}_{q_\phi(\mathbf{z} \mid \mathbf{x}^{local})}
\big[ \log p_\theta(\boldsymbol{x}^{(j)} \mid \mathbf{z}) \big] \\
&- 
\mathbb{E}_{p_{\mathcal{D}}(\mathbf{x})}
\mathrm{KL}\!\left(q_\phi(\mathbf{z} \mid \mathbf{x}^{local}) \,\|\, p(\mathbf{z})\right), \\
\end{aligned}
\label{Eq:unidwm0}
\end{equation}
where $z \in \mathcal{Z}$ denotes the latent variable, $p_{\theta}(\boldsymbol{x}^{(j)}|\mathbf{z})$ represents single-modality conditional distribution, and $q_{\phi}(z | \mathbf{x}^{local})$ is the approxiation posterior given local observations. A proof can be found in Appendix~\ref{vae_unidwm}. However, the VAE-style ELBO may fail to learn meaningful latent representation, especially when the data space $\mathcal{X}$ is complex and high-dimensional. Therefore, we conclude another ELBO objective inspired by InfoVAE:
\begin{equation}
\begin{aligned}
\mathcal{L}_{\text{UniDWM}} =&
 {\textstyle \sum_{j=1}^{M}} \mathbb{E}_{p_{\mathcal{D}}(\mathbf{x})}
\mathbb{E}_{q_\phi(\mathbf{z} \mid \mathbf{x}^{local})}
\big[ \log p_\theta(\boldsymbol{x}^{(j)} \mid \mathbf{z}) \big] \\
&- \lambda D(q_{\phi}(\mathbf{z})\,\|\,p(\mathbf{z})), \\ 
\end{aligned}
\label{Eq:unidwm}
\end{equation}
which is similar to Eq.~\ref{Eq:infovae}, but with multi-observation reconstruction and setting $a=1$. A proof can be seen in Appendix~\ref{infovae_unidwm}. In practice, we employ the SIGReg \cite{balestriero2025lejepa} as the regulation term $D(\cdot\|\cdot)$. SIGReg applies the Epps-Pulley test to measure the distribution discrepancy, which satisfies the requirement of InfoVAE ($D(q_\phi(z) \|\ p(z)) = 0$ iff $q_\phi(z) = p(z)$). The log-likelihood terms are implemented as the negative reconstruction and generation losses $\mathcal{L}_{recon}$ and $\mathcal{L}_{gen}$, which will be detailed in Sec.~\ref{recon} and Sec.~\ref{gen}.

\subsection{Framework of UniDWM}
The overall framework of UniDWM is illustrated in Figure~\ref{fig_framework}. UniDWM adopts a spatiotemporal encoder–decoder architecture: the encoder compresses visual inputs into a compact latent feature field, while a family of decoders reconstructs different physical aspects of the driving scene from this representation. To enable such comprehensive representation learning, UniDWM first introduces \textbf{\textit{Joint Reconstruction}} (Sec.~\ref{recon}), which guides the encoder be aware of spatial structure by jointly reconstructing geometry, appearance, and ego motion from local observations. Subsequently, the pretrained model is fine-tuned in the \textbf{\textit{Collaborative Generation}} stage (Sec.~\ref{gen}) to further encode dynamic cues into the latent world representation. Together, these components allow UniDWM to function not only as a self-supervised foundation model for world modeling but also as a unified framework supporting a wide range of downstream tasks, including reasoning about geometric and visual structures (perception), modeling the dynamics of surrounding agents (prediction), and capturing ego-motion evolution (planning).


\subsection{Joint Reconstruction}
\label{recon}
To learn a physically grounded latent world representation, we construct the world representation through a joint 4D scene reconstruction pathway that reconstructs 3D geometry, visual appearance, and ego motion over time. Specifically, we adopt a variational autoencoding architecture, where a shared spatiotemporal encoder $q_\phi$ compresses a sequence of visual observations $I_{1:n}$ and ego status $s_{1:n}$ into a unified latent embedding $\mathbf{z}$:
\begin{equation}
    \mathbf{z} \sim q_\phi(\mathbf{z}\mid I_{1:n}, s_{1:n}),
\end{equation}
where $\mathbf{z}$ is designed to be \textit{modality-agnostic} and \textit{time-continuous}, enabling it to jointly encode 3D structural priors, temporal motion cues, and visual appearance information within a shared latent space. The shared spatiotemporal encoder consists of two complementary components: a static encoder and a dynamic encoder. 

\noindent\textbf{Static Encoder.} To reduce training overhead and stabilize the representation learning, we reuse a pretrained encoder \cite{dcae, dinov3} as the static encoder, which is responsible for capturing view-consistent features from individual frames. The parameters of this static encoder are frozen during training to preserve general low-level representations. The ego status is encoded by an MLP. A deep-spatial-compressed latent state $\textbf{z}\in \mathbb{R}^{n\times (l+1) \times c}$ is obtained:
\begin{equation}
    \begin{aligned}
     \textbf{z}_{rgb} = \text{Encoder}(I_{1:n}) \in \mathbb{R}^{n\times l \times c}, \\
     \textbf{z}_{ego} = \text{MLP}(s_{1:n}) \in \mathbb{R}^{n\times 1 \times c}, \\
     \textbf{z} = \text{Concatenate}([\textbf{z}_{rgb},\textbf{z}_{ego}]),
    \end{aligned}
    \label{Eq:s_encoder}
\end{equation}
\noindent where $c$ is the channel number, $l$ is the length of tokens of each frame.

\begin{figure}[t]
    \centering
    \includegraphics[width=1.0\columnwidth]{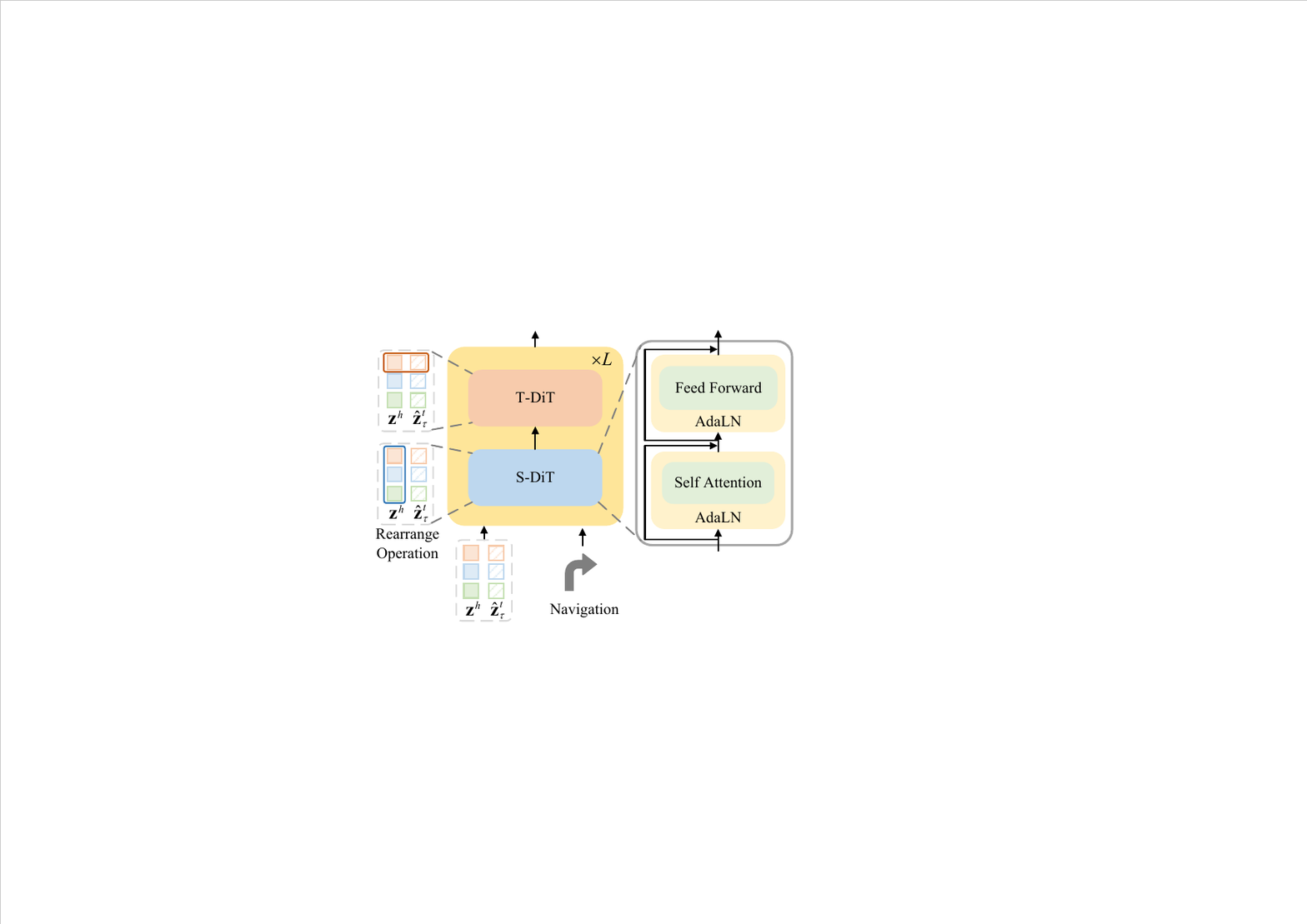}
    \caption{\textbf{Architecture of the DiT block}, consisting of an alternating sequence of spatial DiT (S-DiT) and temporal DiT (T-DiT) modules. }
    \label{fig_causal}
\end{figure}


\noindent\textbf{Dynamic Encoder.} The dynamic encoder is designed to model scene evolution and ego dynamics over time. Inspired by recent advances in spatiotemporal and geometry modeling~\cite{hu2024drivingworld, epona, vggt}, we construct the dynamic encoder by alternately stacking spatial and temporal attention layers. The architecture of the dynamic encoder is shown in Figure~\ref{fig_framework}. The spatial attention aggregates geometric and contextual features within each frame, while the temporal attention captures inter-frame motion and continuity in the latent domain. This alternating structure enables effective cross-frame interaction and smooth temporal reasoning, ensuring that the resulting latent representation $\mathbf{z}$ reflects a consistent 4D understanding of the driving scene. Specifically, a stack of $L$ layers of spatial-temporal transformers is employed to capture the structure information of the driving scene progressively:
\begin{equation}
    \begin{aligned}
     \textbf{z} & \gets  \text{SpatialAttn}(\textbf{z}), \\
     \textbf{z} &\gets  \text{TemporalAttn}(\textbf{z},
     \mathcal{M}),
    \end{aligned}
    \label{Eq:d_encoder}
\end{equation}
\noindent where $\mathcal{M}\in \mathbb{R}^{n \times n}$ is a causal mask that is used to keep information flowing unidirectionally over time. 

\noindent\textbf{Decoupled Decoders.}
Given the shared latent world representation $\mathbf{z}$, the objective is to reconstruct geometry, visual appearance, and ego motion, respectively. Since these modalities reside in distinct feature domains, we design decoupled decoders that project $\textbf{z}$ into their respective feature spaces:
\begin{equation}
\begin{aligned}
    \{\hat{\textbf{D}}, \hat{\textbf{P}},\Sigma_{d},\Sigma_{p}  \} =\mathcal{D}^{\text{geo}}(\textbf{z}), \\
    \hat{\textbf{C}} = \mathcal{D}^{\text{rgb}}(\textbf{z}), \hat{\textbf{e}}=\mathcal{D}^{\text{ego}}(\textbf{z}),
\end{aligned}
\end{equation}
where $\{\hat{\textbf{D}} \in \mathbb{R}^{n\times H \times W \times 3},\hat{\textbf{P}} \in \mathbb{R}^{n\times H \times W \times 1}\}$ denote the reconstructed geometry attributes (points and depths). $\{\Sigma_d, \Sigma_p\}$ represent the aleatoric uncertainty \cite{kendall2016modelling, novotny2018capturing} of depth and point predictions. $\hat{\textbf{C}} \in \mathbb{R}^{n \times H \times W \times 3}$, and $\hat{\textbf{e}} \in \mathbb{R}^{n \times 3}$ denote the visual appearance (RGB colors) and ego poses (ego center and heading), respectively. 

For geometry and appearance reconstruction, we build decoders following prior methods \cite{dcae, rae}. In practice, $\mathcal{D}^{\text{rgb}}$ are initialized from the pretrained decoder weights \cite{dcae, rae} and frozen during training. For the ego-motion decoder, following VGGT~\cite{vggt}, we employ a lightweight transformer-based decoder composed of four self-attention layers followed by a linear projection head.

\noindent\textbf{Reconstruction Loss. }
We employ three kinds of loss to build the reconstruction loss:
\begin{equation}
   \mathcal{L}_{recon} = \mathcal{L}_{ego} + \mathcal{L}_{geo} + \mathcal{L}_{vis},
\end{equation}
\noindent where $\mathcal{L}_{ego}$, $\mathcal{L}_{geo}$, and $\mathcal{L}_{vis}$ are ego motion, geometry, and visual appearance losses, respectively. $\mathcal{L}_{ego}$ is calculated by comparing the L2 distance between the predicted ego poses $\hat{\textbf{e}}$ and ground-truth ego poses \textbf{e}: $\mathcal{L}_{ego} = \frac{1}{n}||\hat{\textbf{e}} - \textbf{e}||_2$. Following prior work \cite{vggt}, $\mathcal{L}_{geo}$ is calculated as follows:
\begin{equation}
\begin{aligned}
    \mathcal{L}_{geo} =& \mathcal{L}_{\text {depth }}  + \mathcal{L}_{\text {point }},\\
    \mathcal{L}_{\text {depth }} =&\frac{1}{n} \| \Sigma_{d} \left(\hat{\textbf{D}}-\textbf{D}\right)\|+ \\ &\frac{1}{n}\|\Sigma_{d} \odot\left(\nabla \hat{\textbf{D}}-\nabla \textbf{D}\right) \|-a \log \Sigma_{d},\\
    \mathcal{L}_{\text {points}} =&\frac{1}{n} \| \Sigma_{p} \left(\hat{\textbf{P}}-\textbf{P}\right)\|+ \\ &\frac{1}{n}\|\Sigma_{p} \odot\left(\nabla \hat{\textbf{P}}-\nabla \textbf{P}\right) \|- a \log \Sigma_{p},\\
\end{aligned}
\end{equation}
\noindent where $\odot$ denotes the channel-broadcast element-wise product, $\Sigma_{d}$ and $\Sigma_{p}$ represent the depth and point uncertainty map, respectively. $\mathcal{L}_{rgb}$ is calculated as follows:
\begin{equation}
\begin{aligned}
    \mathcal{L}_{vis} =& \|\hat{\textbf{C}} - \textbf{C}\|  + \omega_{l}\text{LPIPS}(\hat{\textbf{C}}, \textbf{C}) + \omega_{g} \mathcal{L}_{GAN},\\
\end{aligned}
\end{equation}
\noindent where $\text{LPIPS}(\cdot)$ denotes the LPIPS \cite{lpips} function, $\mathcal{L}_{GAN}$ is GAN loss as in \cite{gan_loss, rae}. $\omega_l$ and $\omega_g$ are weights for LPIPS and GAN loss and are set to $1.0$ and $0.75$ in practice following \cite{rae}.

\subsection{Collaborative Generation}
\label{gen}

While joint 4D scene reconstruction enables the model to learn a physically grounded latent representation of the observed scenes, autonomous driving requires not only understanding but also \emph{forecasting} future dynamics. To achieve such a goal, we introduce a \textit{collaborative generation} framework that models future world evolution as a conditional generative process in the latent space. By jointly sampling future 4D scene latents conditioned on historical context, the model achieves coherent forecasting across different tasks, enabling unified imagination of both future scenes and agent behaviors. Note that generation modules can also be seen as decoders for future observations that enable UniDWM with dynamic imagination capability. 

Specifically, given the latent world representation $\mathbf{z}^{h}\in \mathbb{R}^{n^h \times (l + 1) \times c}$ encoded from historical visual observations and ego poses, our diffusion-based generator aims to synthesize the future latent world representation $\mathbf{z}^{t} \in \mathbb{R}^{n^t \times (l + 1) \times c}$ for 4D generation and planning. Following prior works \cite{kong2024hunyuanvideo, epona}, the generation process is implemented using a conditional DiT framework; the detailed architecture is shown in Figure~\ref{fig_causal}. The DiT framework is also composed of spatial and temporal attention layers similar to the dynamic encoder, but with AdaLN layers to adapt to the timestep and guidance information.

\begin{table*}[t]
\centering
\caption{\textbf{Performance on the NAVSIM \textit{navtest} split.} `*' denotes the results obtained by ourselves. `Percep. Label' denotes the required perception labels for a method. `C', `L', and `E' represent camera, LiDAR, and ego status inputs, respectively. The best results of methods with and without perception labels are highlighted in \textbf{bold}, respectively. }
\begin{tabular}{lccccccc}
\toprule
\textbf{Method} & \textbf{Input} & \textbf{Percep. Label} & \textbf{NC} $\uparrow$ & \textbf{DAC} $\uparrow$ & \textbf{EP} $\uparrow$ & \textbf{TTC} $\uparrow$ & \textbf{PDMS} $\uparrow$ \\
\midrule
UniAD \cite{hu2023planning} & C & BBox, Map, Occ & 97.8 & 91.9 & 78.8 & 92.9 & 83.4 \\
LTF \cite{transfuser} & C & BBox, Map & 97.4 & 92.8 & 79.0 & 92.4 & 83.8 \\
TransFuser \cite{transfuser} & C \& L & BBox, Map & 97.7 & 92.8 & 79.2 & 93.0 & 84.0 \\
Hydra-MDP \cite{li2024hydra} & C \& L & BBox, Map & 98.3 & 96.0 & 78.7 & 94.6 & 86.5 \\
DiffusionDrive \cite{liao2025diffusiondrive} & C \& L & BBox, Map & 98.2 & 96.2 & 82.2 & 94.7 & 88.1 \\
GoalFlow \cite{xing2025goalflow} & C \& L & BBox, Map & 98.4 & \textbf{98.3} & 85.0 & 94.6 & 90.3 \\
GaussianFusion \cite{liu2025gaussianfusion} & C \& L & BBox, Map & \textbf{98.7} & 98.1 & \textbf{88.2} & \textbf{95.7} & \textbf{92.0} \\
\hline
Ego-MLP & E & None & 93.0 & 77.3 & 62.8 & 83.6 & 65.6 \\
LAW \cite{law} & C & None & 97.2 & 93.3 & 78.8 & 91.9 & 83.8 \\
World4Drive \cite{zheng2025world4drive} & C & None & 97.4 & 94.3 & 79.9 & 92.8 & 85.1 \\
DINOv3 (B)$^*$ \cite{dinov3} & C & None & 97.8 & 94.4 & 80.2 & 93.0 & 85.4 \\
Epona \cite{epona} & C & None & 97.9 & 95.1 & 80.4 & 93.8 & 86.2 \\
\rowcolor{cyan!20}
UniDWM (DCAE) & C & None & 97.4 & 93.2 & 80.1 & 93.3 & 84.9 \\
\rowcolor{cyan!20}
UniDWM (DINOv3-B) & C & None & \textbf{98.3} & \textbf{97.9} & \textbf{86.0} & \textbf{95.1} & \textbf{90.6} \\
\bottomrule
\end{tabular}
\label{tab:navsim_pdms}
\end{table*}



In practice, we inject navigation embedding $\mathbf{g}_{\text{nav}}$ as conditional guidance to the DiT blocks. This enables the generator to jointly imagine plausible future latent state $\hat{\mathbf{z}}^{t}$ that are semantically aligned with navigation intent. The conditional generative dynamics are formulated as:
\begin{equation}
    \hat{\mathbf{z}}^{t}_{\tau-1} = 
    \hat{\mathbf{z}}^{t}_{\tau} + \Delta \tau \cdot v_{\Theta}\left(\hat{\mathbf{z}}^{t}_{\tau}, \tau, [\mathbf{z}^{h}, \mathbf{g}_{\text{nav}}]\right),
    \label{eq:stage1}
\end{equation}
where $\hat{\mathbf{z}}^{t}_{\tau}$ denotes the noisy future latent at diffusion timestep $\tau$, $\mathbf{z}^{h}$ represents the historical latent context, and $v_{\Theta}$ is the DiT-based velocity field. The loss for the generation follows the standard velocity prediction objective:
\begin{equation}
    \mathcal{L}_{\text{gen}} = 
    \mathbb{E}_{\mathbf{z}^{t}, \epsilon, \tau}
    \left\|v_{\Theta}\left(\hat{\mathbf{z}}^{t}_{\tau}, \tau, [\mathbf{z}^{h}, \mathbf{g}_{\text{nav}}]\right)
    - (\mathbf{z}^{t} - \epsilon)\right\|^{2},
    \label{eq:loss1}
\end{equation}
where $\mathbf{z}^{t}$ denotes the ground-truth future latent representation, and $\epsilon \sim \mathcal{N}(0, I)$.

%% file: sec/4_exp.tex
\section{Experiment}
\begin{figure*}[t]
    \centering
    \includegraphics[width=0.98\textwidth]{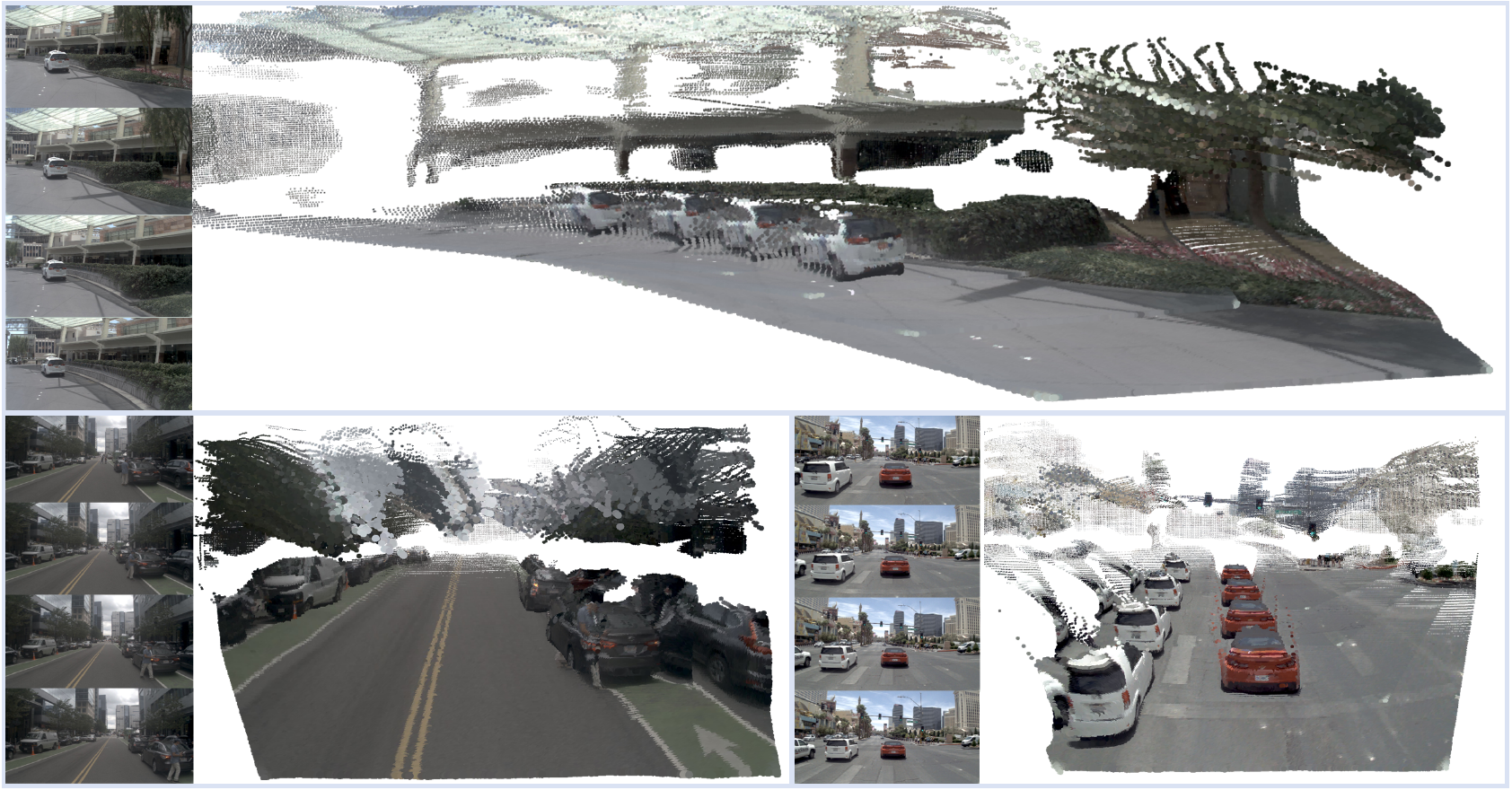}
    \caption{\textbf{Visualization of 4D recontruction results.} The left images are the sequential visual inputs, the right visualized points are the reconstructed geometry. }
    \label{fig_recon_0}
\end{figure*}

\begin{figure*}[t]
    \centering
    \includegraphics[width=1.0\textwidth]{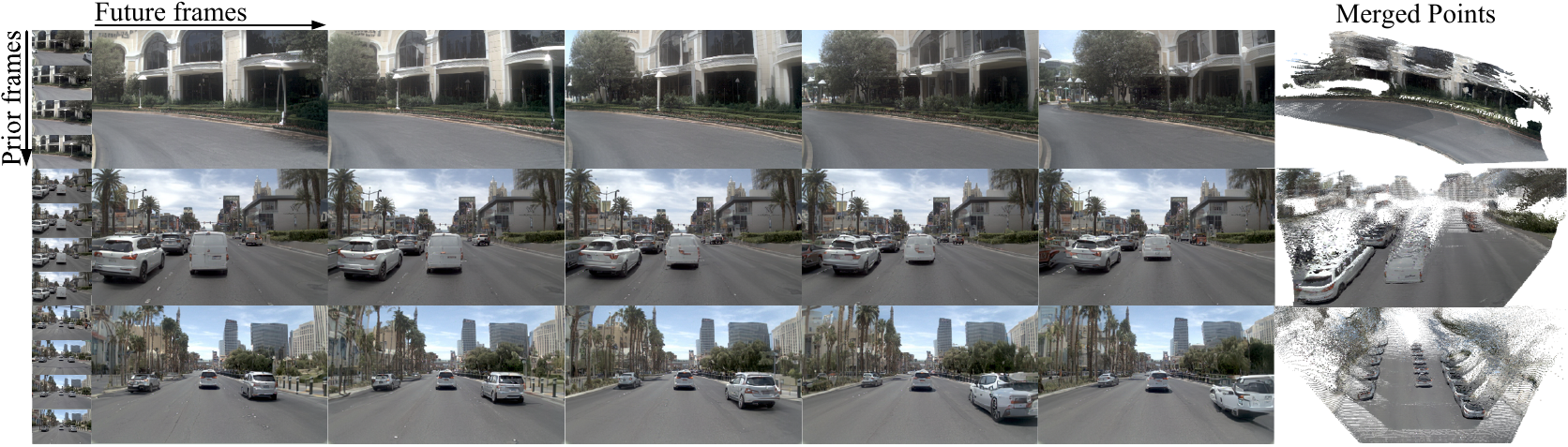}
    \caption{\textbf{Visualization of 4D generation results.} Given sequential prior visual frames, UniDWM generates future visual frames with corresponding geometry. We merge the geometries and visualize them as colored points. }
    \label{fig_gen_0}
\end{figure*}

We primarily evaluate our models on the NAVSIM~\cite{navsim} dataset. Details of the dataset and evaluation metrics are provided in Appendix~\ref{dataset_metrics}.

\subsection{Implementation Details}
We utilize DINOv3-B \cite{dinov3} and DCAE \cite{dcae} as the static encoder, separately. The dynamic encoder is composed of $12$-layer spatiotemporal transformers with $170$M parameters. The reconstruction decoders for geometry, appearance, and ego pose are independent and built following \cite{vggt}. We employ a $12$-layer next-frame prediction diffusion transformer with $459$M parameters, and a $16$-layer trajectory prediction diffusion transformer with $33$M parameters. The point and depth map annotations are obtained by projecting the LiDAR points to the image plane. The training recipe is composed of two stages: 1) Reconstruction training, where the model is trained from scratch on the \textit{navtrain} split of the NAVSIM dataset for reconstruction objective $\mathcal{L}_{recon} + \lambda \text{SIGReg}$; 2) Reconstruction and generation joint training, where the model inherits the weights from the first stage training and keeps training with objective $\mathcal{L}_{gen} + \mathcal{L}_{recon} + \lambda \text{SIGReg}$. $\lambda$ was set to $2 \times10^{-4}$ for all experiments. All input images are resized to a resolution of $224 \times 384$ pixels. 50 epochs were performed with a batch size of $64$ for both the first and second stages. The AdamW optimizer was utilized, with a learning rate set to $1 \times 10^{-4}$ and a weight decay of $5 \times 10^{-2}$. In all experiments, the DiT sampling step for the next frame and trajectory were configured to $100$ and $5$, respectively.


\subsection{Experimental Results}

\noindent\textbf{Evaluation of Trajectory Planning.}
We evaluate UniDWM on the NAVSIM \textit{navtest} split and compare it with a wide range of representative trajectory planning methods, including both perception-label-supervised and label-free approaches. As shown in Table~\ref{tab:navsim_pdms}, methods that leverage dense perception annotations generally achieve strong performance, with GaussianFusion~\cite{liu2025gaussianfusion} attaining the best overall results among perception-supervised methods. Focusing on label-free settings, UniDWM demonstrates competitive performance without relying on any perception labels. In particular, UniDWM with a DINOv3-B backbone achieves the best results among all label-free methods, significantly outperforming prior approaches such as raw DINOv3 \cite{dinov3}, Epona~\cite{epona}, and World4Drive~\cite{zheng2025world4drive} across all evaluation metrics. Moreover, comparing UniDWM variants with different encoders highlights the importance of a static encoder with sufficient capacity. Replacing the DCAE encoder with DINOv3-B yields substantial improvements across all metrics, demonstrating that UniDWM can effectively leverage stronger encoders. Overall, these results suggest that UniDWM provides a scalable and annotation-efficient alternative for trajectory planning, narrowing the performance gap between perception label-free and fully supervised methods.

\begin{table}[t]
\centering
\caption{\textbf{4D reconstruction results on the NAVSIM \textit{navtest} split.}}
\begin{tabular}{cccc}
    \toprule
    \textbf{Method} & \textbf{Acc.$\downarrow$} & \textbf{Comp.$\downarrow$} & \textbf{Overall$\downarrow$}\\
    \midrule
    VGGT & 3.764 & 2.236 & 3.000 \\
    Spann3R & 2.028 & 2.202 & 2.115  \\
    UniDWM (ours) & \textbf{1.878} & \textbf{1.575} & \textbf{1.727} \\
    \bottomrule
\end{tabular}
\label{tab:3r}
\end{table}

\begin{table*}[t]
\centering
\caption{\textbf{Ablation study} of different architecture configurations. Results are obtained on the NAVSIM \textit{navtest} split. }
\begin{tabular}{cccc|ccccc}
\toprule
\multirow{2}{*}{\textbf{\textit{\shortstack{Ego Pose\\ Reconstruction}}}} & \multirow{2}{*}{\textbf{\textit{\shortstack{Appearance\\ Reconstruction}}}} & \multirow{2}{*}{\textbf{\textit{\shortstack{Geometry \\Reconstruction}}}} & \multirow{2}{*}{\textbf{\textit{\shortstack{Dynamic \\Generation}}}} &\multirow{2}{*}{\textbf{NC$\uparrow$}} & \multirow{2}{*}{\textbf{DAC$\uparrow$}} & \multirow{2}{*}{\textbf{EP$\uparrow$}} & \multirow{2}{*}{\textbf{TTC$\uparrow$}} & \multirow{2}{*}{\textbf{PDMS$\uparrow$}} \\
&&&&&&&&\\
\midrule
\ding{51} & \ding{55} & \ding{55} & \ding{55} & 96.5 & 88.3 & 74.3 & 90.5 & 78.5 \\
\ding{51} & \ding{51} & \ding{55} & \ding{55} & 97.6 & 90.1 & 77.1 & 91.6 & 81.3 \\
\ding{51} & \ding{55} & \ding{51} & \ding{55} & 96.8 & 90.7 & 76.3 & 91.1 & 80.0 \\
\ding{51} & \ding{55} & \ding{55} & \ding{51} & 96.8 & 90.7 & 76.3 & 91.1 & 80.9 \\
\ding{51} & \ding{55} & \ding{51} & \ding{51} & 97.3 & 90.7 & 76.8 & 91.7 & 81.6 \\
\rowcolor{cyan!20}
\ding{51} & \ding{51} & \ding{51} & \ding{51} & 97.3 & 91.6 & 77.4 & 92.1 & 82.4 \\
\bottomrule
\end{tabular}
\label{tab:ablation}
\end{table*}

\begin{table}[t]
\centering
\caption{\textbf{Additional finetune with GRPO.} The model in the first row of Table~\ref{tab:ablation} is used as the baseline.}
\resizebox{\columnwidth}{!}{
\begin{tabular}{c|ccccc}
\toprule
\textbf{Method} & \textbf{NC$\uparrow$} & \textbf{DAC$\uparrow$} & \textbf{EP$\uparrow$} & \textbf{TTC$\uparrow$} & \textbf{PDMS$\uparrow$} \\
\midrule
Baseline            & 96.5 & 88.3 & 74.3 & 90.5 & 78.5 \\
Baseline w/ GRPO    & 96.6 & 90.4 & 76.6 & 91.8 & 81.2 \\
UniDWM              & 97.3 & 91.6 & 77.4 & 92.1 & 82.4 \\
UniDWM w/ GRPO      & \textbf{97.4} & \textbf{93.2} & \textbf{80.1} & \textbf{93.3} & \textbf{84.9} \\
\bottomrule
\end{tabular}
}
\label{tab:grpo}
\end{table}

\noindent\textbf{Evaluation of 4D Reconstruction.}
To evaluate the 4D reconstruction capability of UniDWM, we compare it with VGGT~\cite{vggt} and Spann3R~\cite{wang20253d} on the NAVSIM \textit{navtest} split, using Accuracy ($\text{Acc.}$), Completeness ($\text{Comp.}$), and Overall Chamfer Distance ($\text{Overall}$) as metrics. As shown in Table~\ref{tab:3r}, UniDWM achieves the best performance across all metrics, with the lowest Overall ($1.727$), Accuracy ($1.878$), and Completeness ($1.575$). Benefiting from the joint 4D reconstruction (Sec.~\ref{recon}), UniDWM reduces the Overall metric by $\mathbf{42.4\%}$ compared to VGGT, demonstrating its effectiveness in producing high-fidelity and temporally coherent 4D reconstructions. Qualitative results in Figure~\ref{fig_recon_0} further validate these findings. UniDWM reconstructs fine-grained geometric details (e.g., overhead structures and ground markings), maintains strong temporal coherence for dynamic objects (e.g., vehicles), and achieves good scene completeness (e.g., traffic lights). Together, these results confirm the robustness and effectiveness of UniDWM for 4D reconstruction.

\noindent\textbf{Evaluation of 4D Generation.}
Figure~\ref{fig_gen_0} shows the 4D generation results of UniDWM using autoregressive next-frame prediction of both images and point clouds. UniDWM accurately captures the geometry and appearance of static scene elements relative to the ego vehicle, with good alignment between visual content and generated geometry. For dynamic agents, the model preserves reasonable geometric consistency, though visual artifacts may appear over long horizons due to error accumulation from autoregressive inference. This issue arises from the training–inference mismatch and could be mitigated by techniques such as chain-of-forward training \cite{epona}. However, since the primary focus of this work is on the representation learning of driving scenes rather than generation quality optimization, we leave exploration of such techniques to future work.

\subsection{Ablation Study}
To understand the impact of each design choice in UniDWM, we perform a controlled ablation study on the trajectory planning task by incrementally {\textit{Ego Pose Reconstruction}}, {\textit{Appearance Reconstruction}}, {\textit{Geometry Reconstruction}}, and {\textit{Dynamic Generation}} decoders with a shared DCAE encoder \cite{dcae}. The results are shown in Table~\ref{tab:ablation}. When only using the ego pose reconstruction, the PDMS is 78.5. Introducing {\textit{Appearance Reconstruction}}, {\textit{Geometry Reconstruction}}, and {\textit{Dynamic Generation}} leads to +2.8, +1.5, and +2.4 gains in PDMS, respectively. Combining {\textit{Geometry Reconstruction}} and {\textit{Dynamic Generation}} further improves the performance to $81.6$ PDMS. The best overall performance is achieved when combining all reconstruction and generation decoders. These findings confirm that our multifaceted representation learning strategy effectively captures complementary aspects of the driving scene, where appearance, geometry, and dynamics provide mutually reinforcing supervision signals. 

To further explore the useful planning-oriented information encoded in the latent world representation, we freeze the UniDWM encoder and finetune the trajectory decoder using the GRPO algorithm \cite{li2025recogdrive}. As reported in Table~\ref{tab:grpo}, UniDWM without GRPO already achieves competitive or superior performance compared to the GRPO-enhanced baseline. When GRPO is applied to UniDWM, the model achieves the best overall results, suggesting that the benefits of unified latent world modeling and policy optimization are complementary rather than redundant. These results highlight that high-quality latent world modeling not only improves planning performance directly but also enables effective reinforcement-based finetuning. Additional smoothness analysis of representations is in Appendix~\ref{sec:appendix_smoothness}.

%% file: main.bbl
\begin{thebibliography}{47}
\providecommand{\natexlab}[1]{#1}
\providecommand{\url}[1]{\texttt{#1}}
\expandafter\ifx\csname urlstyle\endcsname\relax
  \providecommand{\doi}[1]{doi: #1}\else
  \providecommand{\doi}{doi: \begingroup \urlstyle{rm}\Url}\fi

\bibitem[Balestriero \& LeCun(2025)Balestriero and LeCun]{balestriero2025lejepa}
Balestriero, R. and LeCun, Y.
\newblock Lejepa: Provable and scalable self-supervised learning without the heuristics.
\newblock \emph{arXiv preprint arXiv:2511.08544}, 2025.

\bibitem[Bojarski et~al.(2016)Bojarski, Del~Testa, Dworakowski, Firner, Flepp, Goyal, Jackel, Monfort, Muller, Zhang, et~al.]{bojarski2016end}
Bojarski, M., Del~Testa, D., Dworakowski, D., Firner, B., Flepp, B., Goyal, P., Jackel, L.~D., Monfort, M., Muller, U., Zhang, J., et~al.
\newblock End to end learning for self-driving cars.
\newblock \emph{arXiv preprint arXiv:1604.07316}, 2016.

\bibitem[Chen et~al.(2025)Chen, Cai, Chen, Xie, Yang, Tang, Li, Lu, and Han]{dcae}
Chen, J., Cai, H., Chen, J., Xie, E., Yang, S., Tang, H., Li, M., Lu, Y., and Han, S.
\newblock Deep compression autoencoder for efficient high-resolution diffusion models.
\newblock In \emph{ICLR}, 2025.

\bibitem[Chen et~al.(2024)Chen, Jiang, Gao, Liao, Xu, Zhang, Huang, Liu, and Wang]{chen2024vadv2}
Chen, S., Jiang, B., Gao, H., Liao, B., Xu, Q., Zhang, Q., Huang, C., Liu, W., and Wang, X.
\newblock Vadv2: End-to-end vectorized autonomous driving via probabilistic planning.
\newblock \emph{arXiv preprint arXiv:2402.13243}, 2024.

\bibitem[Codevilla et~al.(2018)Codevilla, M{\"u}ller, L{\'o}pez, Koltun, and Dosovitskiy]{codevilla2018end}
Codevilla, F., M{\"u}ller, M., L{\'o}pez, A., Koltun, V., and Dosovitskiy, A.
\newblock End-to-end driving via conditional imitation learning.
\newblock In \emph{ICRA}, 2018.

\bibitem[Codevilla et~al.(2019)Codevilla, Santana, L{\'o}pez, and Gaidon]{codevilla2019exploring}
Codevilla, F., Santana, E., L{\'o}pez, A.~M., and Gaidon, A.
\newblock Exploring the limitations of behavior cloning for autonomous driving.
\newblock In \emph{ICCV}, 2019.

\bibitem[Dauner et~al.(2024)Dauner, Hallgarten, Li, Weng, Huang, Yang, Li, Gilitschenski, Ivanovic, Pavone, Geiger, and Chitta]{navsim}
Dauner, D., Hallgarten, M., Li, T., Weng, X., Huang, Z., Yang, Z., Li, H., Gilitschenski, I., Ivanovic, B., Pavone, M., Geiger, A., and Chitta, K.
\newblock Navsim: Data-driven non-reactive autonomous vehicle simulation and benchmarking.
\newblock In \emph{NeurIPS}, 2024.

\bibitem[Esser et~al.(2021)Esser, Rombach, and Ommer]{gan_loss}
Esser, P., Rombach, R., and Ommer, B.
\newblock Taming transformers for high-resolution image synthesis.
\newblock In \emph{CVPR}, 2021.

\bibitem[Gao et~al.(2025)Gao, Chen, Xiao, Hong, Li, and Xu]{magicdrivev2}
Gao, R., Chen, K., Xiao, B., Hong, L., Li, Z., and Xu, Q.
\newblock Magicdrive-v2: High-resolution long video generation for autonomous driving with adaptive control.
\newblock In \emph{ICCV}, 2025.

\bibitem[Gao et~al.(2024)Gao, Yang, Chen, Chitta, Qiu, Geiger, Zhang, and Li]{GaoVista2024}
Gao, S., Yang, J., Chen, L., Chitta, K., Qiu, Y., Geiger, A., Zhang, J., and Li, H.
\newblock Vista: A generalizable driving world model with high fidelity and versatile controllability.
\newblock In \emph{NeurIPS}, 2024.

\bibitem[Guo et~al.(2025)Guo, Wu, Xiong, Xu, Zhou, Xu, Xu, Sun, Wang, Chen, et~al.]{guo2025genesis}
Guo, X., Wu, Z., Xiong, K., Xu, Z., Zhou, L., Xu, G., Xu, S., Sun, H., Wang, B., Chen, G., et~al.
\newblock Genesis: Multimodal driving scene generation with spatio-temporal and cross-modal consistency.
\newblock \emph{arXiv preprint arXiv:2506.07497}, 2025.

\bibitem[Hu et~al.(2024)Hu, Yin, Jia, Deng, Guo, Zhang, Long, and Tan]{hu2024drivingworld}
Hu, X., Yin, W., Jia, M., Deng, J., Guo, X., Zhang, Q., Long, X., and Tan, P.
\newblock Drivingworld: Constructing world model for autonomous driving via video gpt.
\newblock \emph{arXiv preprint arXiv:2412.19505}, 2024.

\bibitem[Hu et~al.(2023)Hu, Yang, Chen, Li, Sima, Zhu, Chai, Du, Lin, Wang, et~al.]{hu2023planning}
Hu, Y., Yang, J., Chen, L., Li, K., Sima, C., Zhu, X., Chai, S., Du, S., Lin, T., Wang, W., et~al.
\newblock Planning-oriented autonomous driving.
\newblock In \emph{CVPR}, 2023.

\bibitem[Huang et~al.(2024)Huang, Zheng, Zhang, Zhou, and Lu]{huang2024gaussianformer}
Huang, Y., Zheng, W., Zhang, Y., Zhou, J., and Lu, J.
\newblock Gaussianformer: Scene as gaussians for vision-based 3d semantic occupancy prediction.
\newblock In \emph{ECCV}, 2024.

\bibitem[Jia et~al.(2023)Jia, Wu, Chen, Xie, He, Yan, and Li]{jia2023think}
Jia, X., Wu, P., Chen, L., Xie, J., He, C., Yan, J., and Li, H.
\newblock Think twice before driving: Towards scalable decoders for end-to-end autonomous driving.
\newblock In \emph{CVPR}, 2023.

\bibitem[Jia et~al.(2025)Jia, You, Zhang, and Yan]{jia2025drivetransformer}
Jia, X., You, J., Zhang, Z., and Yan, J.
\newblock Drivetransformer: Unified transformer for scalable end-to-end autonomous driving.
\newblock In \emph{ICLR}, 2025.

\bibitem[Jiang et~al.(2023)Jiang, Chen, Xu, Liao, Chen, Zhou, Zhang, Liu, Huang, and Wang]{jiang2023vad}
Jiang, B., Chen, S., Xu, Q., Liao, B., Chen, J., Zhou, H., Zhang, Q., Liu, W., Huang, C., and Wang, X.
\newblock Vad: Vectorized scene representation for efficient autonomous driving.
\newblock In \emph{ICCV}, 2023.

\bibitem[Kendall \& Cipolla(2016)Kendall and Cipolla]{kendall2016modelling}
Kendall, A. and Cipolla, R.
\newblock Modelling uncertainty in deep learning for camera relocalization.
\newblock In \emph{ICRA}, 2016.

\bibitem[Kerbl et~al.(2023)Kerbl, Kopanas, Leimk{\"u}hler, and Drettakis]{3dgs}
Kerbl, B., Kopanas, G., Leimk{\"u}hler, T., and Drettakis, G.
\newblock 3d gaussian splatting for real-time radiance field rendering.
\newblock \emph{TOG}, 2023.

\bibitem[Kerbl et~al.(2024)Kerbl, Meuleman, Kopanas, Wimmer, Lanvin, and Drettakis]{kerbl2024hierarchical}
Kerbl, B., Meuleman, A., Kopanas, G., Wimmer, M., Lanvin, A., and Drettakis, G.
\newblock A hierarchical 3d gaussian representation for real-time rendering of very large datasets.
\newblock \emph{TOG}, 2024.

\bibitem[Kingma \& Welling(2013)Kingma and Welling]{kingma2014auto}
Kingma, D.~P. and Welling, M.
\newblock Auto-encoding variational bayes.
\newblock \emph{arXiv preprint arXiv:1312.6114}, 2013.

\bibitem[Kong et~al.(2024)Kong, Tian, Zhang, Min, Dai, Zhou, Xiong, Li, Wu, Zhang, et~al.]{kong2024hunyuanvideo}
Kong, W., Tian, Q., Zhang, Z., Min, R., Dai, Z., Zhou, J., Xiong, J., Li, X., Wu, B., Zhang, J., et~al.
\newblock Hunyuanvideo: A systematic framework for large video generative models.
\newblock \emph{arXiv preprint arXiv:2412.03603}, 2024.

\bibitem[Li et~al.(2025{\natexlab{a}})Li, Fan, He, Wang, Chen, Zhang, and Tan]{law}
Li, Y., Fan, L., He, J., Wang, Y., Chen, Y., Zhang, Z., and Tan, T.
\newblock Enhancing end-to-end autonomous driving with latent world model.
\newblock In \emph{ICLR}, 2025{\natexlab{a}}.

\bibitem[Li et~al.(2025{\natexlab{b}})Li, Wang, Liu, He, Fan, and Zhang]{wote}
Li, Y., Wang, Y., Liu, Y., He, J., Fan, L., and Zhang, Z.
\newblock End-to-end driving with online trajectory evaluation via bev world model.
\newblock In \emph{ICCV}, 2025{\natexlab{b}}.

\bibitem[Li et~al.(2025{\natexlab{c}})Li, Xiong, Guo, Li, Yan, Xu, Zhou, Chen, Sun, Wang, et~al.]{li2025recogdrive}
Li, Y., Xiong, K., Guo, X., Li, F., Yan, S., Xu, G., Zhou, L., Chen, L., Sun, H., Wang, B., et~al.
\newblock Recogdrive: A reinforced cognitive framework for end-to-end autonomous driving.
\newblock \emph{arXiv preprint arXiv:2506.08052}, 2025{\natexlab{c}}.

\bibitem[Li et~al.(2024{\natexlab{a}})Li, Li, Wang, Lan, Yu, Ji, Li, Zhu, Kautz, Wu, et~al.]{li2024hydra}
Li, Z., Li, K., Wang, S., Lan, S., Yu, Z., Ji, Y., Li, Z., Zhu, Z., Kautz, J., Wu, Z., et~al.
\newblock Hydra-mdp: End-to-end multimodal planning with multi-target hydra-distillation.
\newblock \emph{arXiv preprint arXiv:2406.06978}, 2024{\natexlab{a}}.

\bibitem[Li et~al.(2024{\natexlab{b}})Li, Wang, Li, Xie, Sima, Lu, Yu, and Dai]{li2024bevformer}
Li, Z., Wang, W., Li, H., Xie, E., Sima, C., Lu, T., Yu, Q., and Dai, J.
\newblock Bevformer: learning bird's-eye-view representation from lidar-camera via spatiotemporal transformers.
\newblock \emph{TPAMI}, 2024{\natexlab{b}}.

\bibitem[Liang et~al.(2022)Liang, Xie, Yu, Xia, Lin, Wang, Tang, Wang, and Tang]{liang2022bevfusion}
Liang, T., Xie, H., Yu, K., Xia, Z., Lin, Z., Wang, Y., Tang, T., Wang, B., and Tang, Z.
\newblock Bevfusion: A simple and robust lidar-camera fusion framework.
\newblock In \emph{NeurIPS}, 2022.

\bibitem[Liao et~al.(2025)Liao, Chen, Yin, Jiang, Wang, Yan, Zhang, Li, Zhang, Zhang, et~al.]{liao2025diffusiondrive}
Liao, B., Chen, S., Yin, H., Jiang, B., Wang, C., Yan, S., Zhang, X., Li, X., Zhang, Y., Zhang, Q., et~al.
\newblock Diffusiondrive: Truncated diffusion model for end-to-end autonomous driving.
\newblock In \emph{CVPR}, 2025.

\bibitem[Liu et~al.(2025)Liu, Liang, Li, Li, and Huang]{liu2025gaussianfusion}
Liu, S., Liang, Q., Li, Z., Li, B., and Huang, K.
\newblock Gaussianfusion: Gaussian-based multi-sensor fusion for end-to-end autonomous driving.
\newblock In \emph{NeurIPS}, 2025.

\bibitem[Liu et~al.(2022)Liu, Wang, Zhang, and Sun]{liu2022petr}
Liu, Y., Wang, T., Zhang, X., and Sun, J.
\newblock Petr: Position embedding transformation for multi-view 3d object detection.
\newblock In \emph{ECCV}, 2022.

\bibitem[Novotny et~al.(2018)Novotny, Larlus, and Vedaldi]{novotny2018capturing}
Novotny, D., Larlus, D., and Vedaldi, A.
\newblock Capturing the geometry of object categories from video supervision.
\newblock \emph{TPAMI}, 2018.

\bibitem[Prakash et~al.(2021)Prakash, Chitta, and Geiger]{transfuser}
Prakash, A., Chitta, K., and Geiger, A.
\newblock Multi-modal fusion transformer for end-to-end autonomous driving.
\newblock In \emph{CVPR}, 2021.

\bibitem[Russell et~al.(2025)Russell, Hu, Bertoni, Fedoseev, Shotton, Arani, and Corrado]{RussellGAIA22025}
Russell, L., Hu, A., Bertoni, L., Fedoseev, G., Shotton, J., Arani, E., and Corrado, G.
\newblock Gaia-2: A controllable multi-view generative world model for autonomous driving.
\newblock \emph{arXiv preprint arXiv:2503.20523}, 2025.

\bibitem[Shi et~al.(2025)Shi, Shi, Sheng, Zhang, and Jiang]{shi2025drivex}
Shi, C., Shi, S., Sheng, K., Zhang, B., and Jiang, L.
\newblock Drivex: Omni scene modeling for learning generalizable world knowledge in autonomous driving.
\newblock In \emph{ICCV}, 2025.

\bibitem[Siméoni et~al.(2025)Siméoni, Vo, Seitzer, Baldassarre, Oquab, Jose, Khalidov, Szafraniec, Yi, Ramamonjisoa, Massa, Haziza, Wehrstedt, Wang, Darcet, Moutakanni, Sentana, Roberts, Vedaldi, Tolan, Brandt, Couprie, Mairal, Jégou, Labatut, and Bojanowski]{dinov3}
Siméoni, O., Vo, H.~V., Seitzer, M., Baldassarre, F., Oquab, M., Jose, C., Khalidov, V., Szafraniec, M., Yi, S., Ramamonjisoa, M., Massa, F., Haziza, D., Wehrstedt, L., Wang, J., Darcet, T., Moutakanni, T., Sentana, L., Roberts, C., Vedaldi, A., Tolan, J., Brandt, J., Couprie, C., Mairal, J., Jégou, H., Labatut, P., and Bojanowski, P.
\newblock Dinov3.
\newblock \emph{arXiv preprint arXiv: 2508.10104}, 2025.

\bibitem[Song et~al.(2025)Song, Jia, Liu, Pan, Zhang, Wang, Zhang, Xu, Yang, and Luo]{song2025don}
Song, Z., Jia, C., Liu, L., Pan, H., Zhang, Y., Wang, J., Zhang, X., Xu, S., Yang, L., and Luo, Y.
\newblock Don't shake the wheel: Momentum-aware planning in end-to-end autonomous driving.
\newblock In \emph{CVPR}, 2025.

\bibitem[Wang \& Agapito(2025)Wang and Agapito]{wang20253d}
Wang, H. and Agapito, L.
\newblock 3d reconstruction with spatial memory.
\newblock In \emph{3DV}, 2025.

\bibitem[Wang et~al.(2025)Wang, Chen, Karaev, Vedaldi, Rupprecht, and Novotny]{vggt}
Wang, J., Chen, M., Karaev, N., Vedaldi, A., Rupprecht, C., and Novotny, D.
\newblock Vggt: Visual geometry grounded transformer.
\newblock In \emph{CVPR}, 2025.

\bibitem[Wang et~al.(2024)Wang, Leroy, Cabon, Chidlovskii, and Revaud]{wang2024dust3r}
Wang, S., Leroy, V., Cabon, Y., Chidlovskii, B., and Revaud, J.
\newblock Dust3r: Geometric 3d vision made easy.
\newblock In \emph{CVPR}, 2024.

\bibitem[Wang et~al.(2022)Wang, Guizilini, Zhang, Wang, Zhao, and Solomon]{wang2022detr3d}
Wang, Y., Guizilini, V.~C., Zhang, T., Wang, Y., Zhao, H., and Solomon, J.
\newblock Detr3d: 3d object detection from multi-view images via 3d-to-2d queries.
\newblock In \emph{CoRL}, 2022.

\bibitem[Xing et~al.(2025)Xing, Zhang, Hu, Jiang, He, Zhang, Long, and Yin]{xing2025goalflow}
Xing, Z., Zhang, X., Hu, Y., Jiang, B., He, T., Zhang, Q., Long, X., and Yin, W.
\newblock Goalflow: Goal-driven flow matching for multimodal trajectories generation in end-to-end autonomous driving.
\newblock In \emph{CVPR}, 2025.

\bibitem[Zhang et~al.(2025)Zhang, Tang, Hu, Pan, Guo, Liu, Huang, Yuan, Zhang, Long, Cao, and Yin]{epona}
Zhang, K., Tang, Z., Hu, X., Pan, X., Guo, X., Liu, Y., Huang, J., Yuan, L., Zhang, Q., Long, X., Cao, X., and Yin, W.
\newblock Epona: Autoregressive diffusion world model for autonomous driving.
\newblock In \emph{ICCV}, 2025.

\bibitem[Zhang et~al.(2018)Zhang, Isola, Efros, Shechtman, and Wang]{lpips}
Zhang, R., Isola, P., Efros, A.~A., Shechtman, E., and Wang, O.
\newblock The unreasonable effectiveness of deep features as a perceptual metric.
\newblock In \emph{CVPR}, 2018.

\bibitem[Zhao et~al.(2017)Zhao, Song, and Ermon]{zhao2017infovae}
Zhao, S., Song, J., and Ermon, S.
\newblock Infovae: Information maximizing variational autoencoders.
\newblock \emph{arXiv preprint arXiv:1706.02262}, 2017.

\bibitem[Zheng et~al.(2025{\natexlab{a}})Zheng, Ma, Tong, and Xie]{rae}
Zheng, B., Ma, N., Tong, S., and Xie, S.
\newblock Diffusion transformers with representation autoencoders.
\newblock \emph{arXiv preprint arXiv:2510.11690}, 2025{\natexlab{a}}.

\bibitem[Zheng et~al.(2025{\natexlab{b}})Zheng, Yang, Xing, Zhang, Zheng, Gao, Li, Zhang, Xia, Jia, et~al.]{zheng2025world4drive}
Zheng, Y., Yang, P., Xing, Z., Zhang, Q., Zheng, Y., Gao, Y., Li, P., Zhang, T., Xia, Z., Jia, P., et~al.
\newblock World4drive: End-to-end autonomous driving via intention-aware physical latent world model.
\newblock In \emph{ICCV}, 2025{\natexlab{b}}.

\end{thebibliography}
